# Deep Predictive Coding Network for Object Recognition

**Haiguang Wen[1], Kuan Han[1], Junxing Shi[1], Yizhen Zhang[1], Eugenio Culurciello[2], Zhongming Liu[1,2]**

## Abstract

Based on the predictive coding theory in neuroscience, we designed a bi-directional and recurrent neural net, namely deep *predictive coding networks (PCN)*. It has feedforward, feedback, and recurrent connections. Feedback connections from a higher layer carry the prediction of its lower-layer representation; feedforward connections carry the prediction errors to its higher-layer. Given image input, PCN runs recursive cycles of bottom-up and top-down computation to update its internal representations and reduce the difference between bottom-up input and top-down prediction at every layer. After multiple cycles of recursive updating, the representation is used for image classification. With benchmark data (CIFAR-10/100, SVHN, and MNIST), PCN was found to always outperform its feedforward-only counterpart: a model without any mechanism for recurrent dynamics. Its performance tended to improve given more cycles of computation over time. In short, PCN reuses a single architecture to recursively run bottom-up and top-down processes. As a dynamical system, PCN can be unfolded to a feedforward model that becomes deeper and deeper over time, while refining it representation towards more accurate and definitive object recognition.

## 1. Introduction

Convolutional neural networks (CNN) have achieved great success in image recognition. Classical CNN models, e.g. AlexNet (Krizhevsky et al., 2012), VGG (Simonyan and Zisserman, 2014), GoogLeNet (Szegedy et al., 2015), ResNet (He et al., 2016b), SENets (Hu et al., 2017), NASNet (Zoph et al., 2017), have improved the performance in computer vision, while these models generally become deeper and wider by using more layers (Simonyan and Zisserman, 2014; Szegedy et al., 2015; He et al., 2016b) or/and filters (Szegedy et al., 2015; Zagoruyko and Komodakis, 2016). Despite various ways of architectural reconfiguration, these models all scale up from the same principle of computation: extracting image features by a feedforward pass through stacks of convolutional layers.

Although it is inspired by hierarchical processing in biological visual systems (Hubel and Wiesel, 1968), CNN differs from the brain in many aspects. Unlike CNN, the brain achieves robust visual perception by using feedforward, feedback and recurrent connections (Felleman and Van, 1991; Sporns and Zwi, 2004). Information is processed not only through a bottom-up pathway running from lower to higher visual areas, but also through a top-down pathway running in the opposite direction. Such bi-directional processes enable humans to perform a wide range of visual tasks, including object recognition. For human vision, feedforward processing is essential to rapid recognition (Serre et al., 2007; DiCarlo et al., 2012), e.g. when visual input is too brief to recruit feedback and recurrent processing (Thorpe et al., 1996). However, feedback processing improves object recognition and enables cognitive processes to influence perception (Logothetis and Sheinberg, 1996; Wyatte et al., 2014). In neuroscience, the interplay between feedforward and feedback processes is described by hierarchical *predictive coding* (Rao and Ballard, 1999; Friston and Kiebel, 2009; George and Hawkins, 2009; Bastos et al., 2012; Clark, 2013; Hohwy, 2013). It states that the feedback connections from a higher visual area to a lower visual area carry predictions of lower-level neural activities; feedforward connections carry the errors between the predictions and the actual lower-level activities. As a result, the brain dynamically updates its representations to progressively refine its perceptual and behavioral decisions.

Based on this brain theory, we designed a bi-directional and recurrent neural net (i.e. PCN). Given image input to PCN, it runs recursive cycles of bottom-up and top-down computation to update its internal representations towards minimization of the residual error between bottom-up input and top-down prediction at every layer in the network. Using predictive coding as its computational mechanism, PCN differs from feedforward-only CNNs that currently dominate computer vision. It is a model with dynamics that uses recursive and bi-directional computation to extract better representations of the input such that the input is predictable by the internal representation. When it is unfolded in time, PCN runs a longer cascade of nonlinear transformations by running more cycles of bottom-up and top-down computation

[1] Electrical and Computer Engineering, Purdue University.
[2] Biomedical Engineering, Purdue University.
Correspondence to: Zhongming Liu < zmliu@purdue.edu>.

through the same architecture without adding more layers, units, or connections.

To explore its value, we designed PCN with convolutional layers stacked in both feedforward and feedback directions. We trained and tested PCN for image classification with benchmark datasets: CIFAR-10 (Krizhevsky and Hinton, 2009), CIFAR-100 (Krizhevsky and Hinton, 2009), SVHN (Netzer et al., 2011), and MNIST (LeCun et al., 1998). Our focus was to explore the intrinsic advantages of PCN over its feedforward-only counterpart: a plain CNN model without feedback connection or any mechanism for recurrent dynamics. It turned out that PCN always outperformed the plain CNN model, and its accuracy tended to improve given more cycles of computation over time. Relative to the classical models, PCN yielded competitive performance in all benchmark tests despite much less layers in PCN. As we did not attempt to optimize the performance by trying many learning parameters or model architectures, there is much room for future studies (e.g. Han et al., 2018) to further improve or extend the model on the basis of a similar notion.

## 2. Related Work

Recent studies demonstrate that deep convolutional neural networks use representations similar to those in the brain (Khaligh-Razavi and Kriegeskorte, 2014; Yamins et al., 2014; Güçlü and van Gerven, 2015; Cichy et al., 2016; Eickenberg et al., 2017; Wen et al., 2017). However, many gaps are yet to be filled to bridge biological and artificial visual systems. A biologically plausible model of vision should take into account feedback and recurrent connections, which are abundant in primate brains (Felleman and Van, 1991; Sporns and Zwi, 2004). A limited number of studies have taken on this direction from the perspective of computational neuroscience or computer vision.

O'Reilly et al. demonstrated that feedback connections could enable top-down representations to fill incomplete bottom-up representations to improve recognition of partially occluded objects (O'Reilly et al., 2013). Exploiting a similar idea, Spoerer et al. built a recurrent CNN (with 2 hidden layers) using feedforward, feedback, and lateral connections to enable recurrent processing that dynamically updated the internal representations as the sum of bottom-up, top-down, and lateral contributions (Spoerer et al., 2017). Trained and tested with synthesized images of digits, their recurrent CNN yielded more robust recognition of digits in cluttered and occluded images. However, that model did not embody an explicit computational mechanism to ensure recurrent processing dynamics to converge over time. Although compelling from the neuroscience perspective, the models in the above studies were relatively simple and shallow, and they were not tested in naturalistic visual scenarios of primary interest to computer vision.

In computer vision, feedback has also played an important role in some vision tasks. For example, feedback was used to select the internal attention to achieve better object recognition performance (Stollenga et al., 2014), or used to model the visual saliency in images (Mahdi and Qin, 2017). Many studies also used a feedback network to reconstruct the visual input in unsupervised learning like autoencoders (Hinton and Salakhutdinov, 2006; Vincent et al., 2010; Masci et al., 2011), deconvolutional networks (Zeiler et al., 2010) and generative models (Hinton, 2012; Canziani and Culurciello, 2017). What remains unresolved is a biologically plausible mechanism that allows feedforward, feedback, and recurrent processes to interact with one another in order for the model to manifest internal dynamics that support various learning objectives.

In this regard, we may seek inspiration from the brain. Predictive coding is an influential theory of neural processing in vision and beyond (Huang and Rao, 2011; Clark, 2013; Hohwy, 2013) as supported by empirical evidence (Gómez et al., 2014; Bastos et al., 2015; Michalareas et al., 2016; Sedley et al., 2016; van Pelt et al., 2016). In a seminal paper (Rao and Ballard, 1997), Rao and Ballard postulated that the brain learns a hierarchical internal model of the visual world. Each level in this model attempts to predict the responses at its lower level via feedback connections; the error between this prediction and the actual response is sent to the higher level via feedforward connections. Friston et al. further generalized this notion into a unified brain theory for perception and action (Friston, 2008). Chalasani et al. used predictive coding to train a deep neural net to learn a hierarchy of sparse representations of data without supervision (Chalasani and Principe, 2013). Lotter et al. explored video prediction as an unsupervised learning objective based on predictive coding (Lotter et al., 2016); however the model trained in this way may not be able to learn sufficiently abstract representation to support such tasks as object recognition. Spratling et al. explored the use of predictive coding for object recognition; however, their model was limited a shallow network architecture for much simplified scenarios (Spratling, 2017).

Inspired by but different from models in prior studies (Rao and Ballard, 1999; Spratling, 2008, 2017), a hierarchical, bidirectional, and recurrent neural network is proposed and implemented herein for object recognition. This model operates with the theory of predictive coding to generate dynamic internal representations by recursive bottom-up and top-down computation. The internal representations are updated to progressively reduce the error of top-down prediction of lower-level representations, while the prediction errors are conveyed upward to higher levels. To train this network, the representations at the highest level, after multiple cycles of recursive updating, are used to classify the input image. With labeled images, the model parameters are trained through backpropagation in time and across layers.

## 3. Methods

### 3.1 Predictive Coding

Central to the theory of predictive coding is that the brain continuously generates top-down predictions of bottom-up inputs. The representation at a higher level predicts the representation at its lower level. The difference between the predicted and actual representation elicits an error of prediction, and propagates to the higher level to update its representation towards improved prediction. This repeats throughout the hierarchy until the errors of prediction diminish, or the bottom-up process no longer conveys any "new" (or unpredicted) information to update the hidden representation. Thus, predictive coding is a computational mechanism for the model to recursively update its internal representations of the visual input towards convergence.

In the following mathematical description of this dynamic process in PCN, italic lowercase letters are used as symbols for *scalars*, bold lowercase letters for column **vectors**, and bold uppercase letters for **MATRICES**. The representation at layer $l$ and time $t$ is denoted as $\mathbf{r}_l(t)$. The weights of feedforward connections from layer $l$-1 to layer $l$ are denoted as $\mathbf{W}_{l-1,l}$. The weights of feedback connections from layer $l$ to layer $l$-1 are denoted as $\mathbf{W}_{l,l-1}$.

In PCN, the higher-level representation, $\mathbf{r}_l(t)$, predicts its lower-level representation as $\mathbf{p}_{l-1}(t)$ via linear weighting $\mathbf{W}_{l,l-1}$, as shown in Eq. (1). The prediction error, $\mathbf{e}_{l-1}(t)$, is the difference between $\mathbf{p}_{l-1}(t)$ and $\mathbf{r}_{l-1}(t)$ as in Eq. (2).

$$\mathbf{p}_{l-1}(t) = \left(\mathbf{W}_{l,l-1}\right)^{\mathrm{T}} \mathbf{r}_l(t) \tag{1}$$

$$\mathbf{e}_{l-1}(t) = \mathbf{r}_{l-1}(t) - \mathbf{p}_{l-1}(\mathrm{t}) \tag{2}$$

### 3.1.1 Feedforward Process

For the feedforward process, the prediction error at layer $l$-1, $\mathbf{e}_{l\text{-}1}(t)$, propagates to the upper layer $l$ to update its representation, $\mathbf{r}_l(t)$, so the updated representation reduces the prediction error. To minimize $\mathbf{e}_{l\text{-}1}(t)$, let's define a loss as the sum of the squared errors normalized by the variance of the representation, $\sigma^2_{l\text{-}1}$, as in Eq. (3).

$$e_{l\text{-}1}(t) = \frac{1}{\sigma^2_{l\text{-}1}} \left\| \mathbf{e}_{l\text{-}1}(t) \right\|_2^2 \tag{3}$$

The gradient of $e_{l\text{-}1}(t)$ with respect to $\mathbf{r}_l(t)$ is as Eq. (4).

$$\frac{\partial e_{l-1}(t)}{\partial \mathbf{r}_l(t)} = -\frac{2}{\sigma^2_{l-1}} \mathbf{W}_{l,l-1} \mathbf{e}_{l-1}(t) \tag{4}$$

To minimize $e_{l\text{-}1}(t)$, $\mathbf{r}_l(t)$ is updated by gradient descent with an updating rate, $\alpha_l$, as shown in Eq. (5).

$$\begin{aligned} \mathbf{r}_l(t+1) &= \mathbf{r}_l(t) - \alpha_l \left(\frac{\partial e_{l-1}(t)}{\partial \mathbf{r}_l(t)}\right) \\ &= \mathbf{r}_l(t) + \frac{2\alpha_l}{\sigma^2_{l-1}} \mathbf{W}_{l,l-1} \mathbf{e}_{l-1}(t) \end{aligned} \tag{5}$$

If the weights of feedback connections are the transpose of those of feedforward connections $\mathbf{W}_{l,l-1} = \left(\mathbf{W}_{l-1,l}\right)^{\mathrm{T}}$, the update rule in Eq. (5) can be rewritten as a *feedforward* operation, as in Eq. (6).

$$\mathbf{r}_l(t+1) = \mathbf{r}_l(t) + a_l \left(\mathbf{W}_{l-1,l}\right)^{\mathrm{T}} \mathbf{e}_{l-1}(t) \tag{6}$$

where the last term indicates forwarding the prediction error from layer $l$-1 to layer $l$ to update the representation with an updating rate $a_l = \frac{2\alpha_l}{\sigma^2_{l-1}}$.

### 3.1.2 Feedback Process

For the feedback process, the top-down prediction is used to update the representation at layer $l$, $\mathbf{r}_l(t)$, to reduce the prediction error $\mathbf{e}_l(t)$. Similar to feedforward process, the error is minimized by gradient descent, where the gradient of $e_l(t)$ with respect to $\mathbf{r}_l(t)$ is as Eq. (7), and $\mathbf{r}_l(t)$ is updated with an updating rate $\beta_l$ as shown in Eq. (8).

$$\frac{\partial e_l(t)}{\partial \mathbf{r}_l(t)} = \frac{2}{\sigma_l^2}\left(\mathbf{r}_l(t) - \mathbf{p}_l(\mathrm{t})\right) \tag{7}$$

$$\begin{aligned} \mathbf{r}_l(t+1) &= \mathbf{r}_l(t) - \beta_l \left(\frac{\partial e_l(t)}{\partial \mathbf{r}_l(t)}\right) \\ &= \left(1 - \frac{2\beta_l}{\sigma_l^2}\right) \mathbf{r}_l(t) + \frac{2\beta_l}{\sigma_l^2} \mathbf{p}_l(t) \end{aligned} \tag{8}$$

Let $b_l = \frac{2\beta_l}{\sigma_l^2}$ and Eq. (8) is rewritten as follows.

$$\mathbf{r}_l(t+1) = (1 - b_l)\mathbf{r}_l(t) + b_l \mathbf{p}_l(t) \tag{9}$$

E. (9) reflects a *feedback* process that the representation at the higher layer, $\mathbf{r}_{l+1}(t)$, generates a top-down prediction, $\mathbf{p}_l(t)$, and influences the lower-layer representation, $\mathbf{r}_l(t)$.

### 3.1.3 Nonlinearity

To add nonlinearity to the above feedforward and feedback processes, a nonlinear activation function is applied to the output of each convolutional layer (except the input layer, i.e. $l = 0$). A rectified linear unit (ReLU) (Nair and Hinton, 2010) converts Eqs. (6) and (9) to nonlinear processes as below.

Nonlinear feedforward process:

$$\mathbf{r}_l(t+1) = \mathrm{ReLU}\left(\mathbf{r}_l(t) + a_l \left(\mathbf{W}_{l-1,l}\right)^{\mathrm{T}} \mathbf{e}_{l-1}(t)\right) \tag{10}$$

Nonlinear feedback process:

$$\mathbf{r}_l(t+1) = \mathrm{ReLU}\left((1 - b_l)\mathbf{r}_l(t) + b_l \mathbf{p}_l(t)\right) \tag{11}$$

## 3.2 Network Architecture

We used the nonlinear feedforward and feedback processes defined in Eq. (10) and (11) as a computational mechanism of predictive coding. We implemented this computational mechanism in several PCNs, all of which included stacked convolutional layers with feedforward, feedback, and recurrent connections as shown in Fig. 1a. These PCNs were trained and tested for object recognition with four benchmark datasets: CIFAR-10, CIFAR-100, SVHN and MNIST. For comparison, several feedforward-only CNNs were built with the same architecture as the feedforward pathway in corresponding PCNs, and were trained and tested with the same datasets. We refer to these feedforward-only CNNs as the plain networks, from which the PCNs were built upon by adding feedback and recurrent connections for dynamic processing.

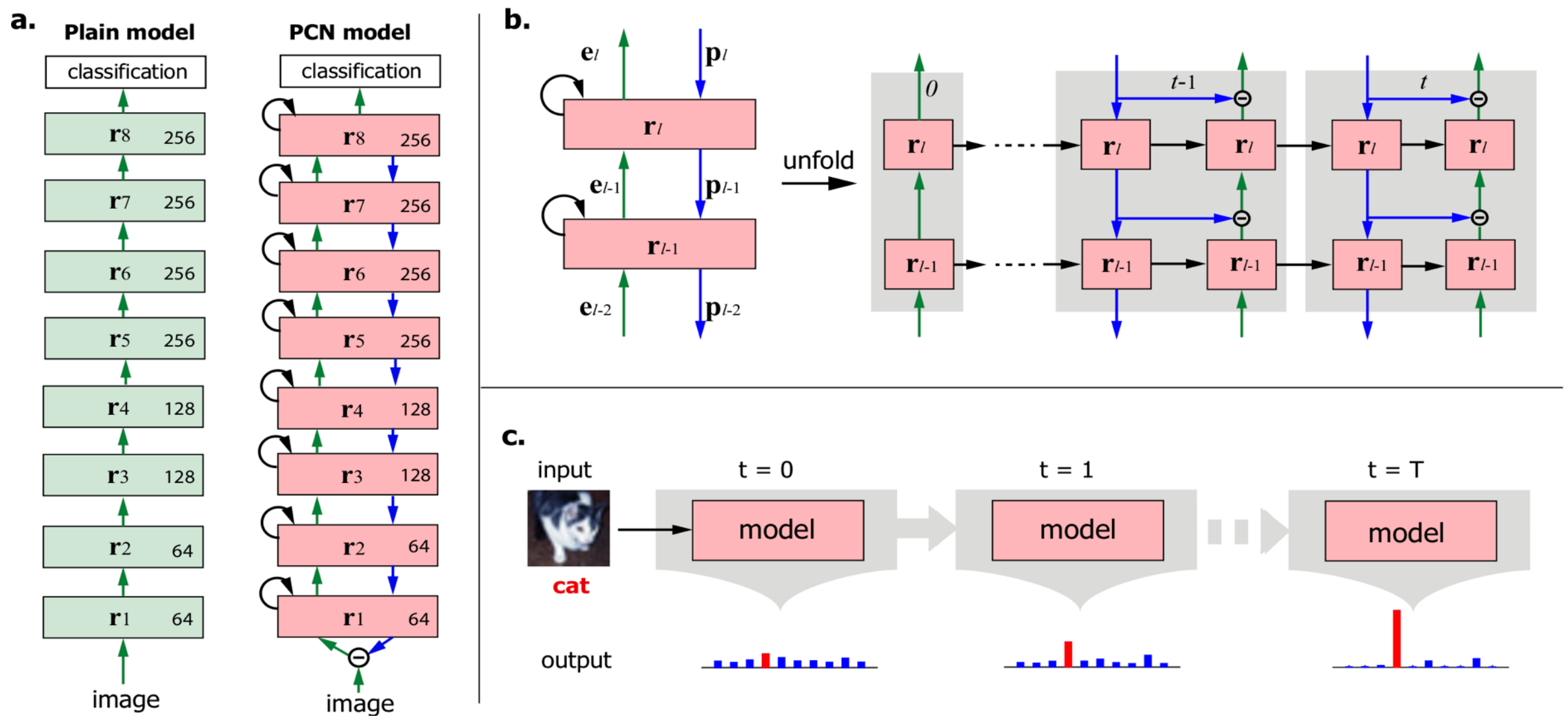

*Figure 1*. a) An example PCN with 9 layers and its CNN counterpart (or the plain model). b) Two-layer substructure of PCN. Feedback (blue), feedforward (green), and recurrent (black) connections convey the top-down prediction, the bottom-up prediction error, and the past information, respectively. c) The dynamic process in the PCN iteratively updates and refines the representation of visual input over time. PCN outputs the probability over candidate categories for object recognition. The bar height indicates the probability and the red indicates the ground truth.

**Plain CNN Models:** The architectural setting of our plain CNN models were similar to the VGG nets (Simonyan and Zisserman, 2014) (see Table 1). Briefly, the basic architecture included 6 or 8 convolutional layers and 1 classification layer. All convolutional layers used 3×3 filters but different numbers of filters, and used rectified linear unit (ReLU) as the nonlinear activation function. For some layers where the number of filters is doubled, the feature maps were reduced by applying 2×2 max-pooling with a stride of 2 after convolution. Batch normalization (Ioffe and Szegedy, 2015) was not used. The classification layer included global average pooling and a fully-connected (FC) layer followed by softmax. On the basis of this setting, we built 5 VGG-like architectures that varied in the number of layers and filters, and trained and tested the models with 4 datasets.

**Predictive Coding Network (PCN):** Starting from each of the plain CNN architectures, we added feedback and recurrent connections to form a corresponding PCN. Fig. 1a shows a 9-layer PCN, running recursive bottom-up and top-down processing based on predictive coding. In PCN, feedback connections from one layer to its lower layer were constrained to be the transposed convolution (Dumoulin and Visin, 2016) which is the transpose of the feedforward counterparts. As such, both feedforward and feedback connections encoded spatial filters. The former was applied to the errors of the top-down prediction of lower-level representation; the latter was applied to high-level representation in order to predict the lower-level representation. As in the brain, feedforward and feedback connections were reciprocal in PCN. The weights of feedback connections had the identical dimension as the transposed weights of feedforward connections. For layers where max-pooling was applied after feedforward convolution, bilinear upsampling was applied before feedback convolution to ensure that the dimension of top-down prediction could match the dimension of lower-level representation.

*Table 1. Architectures for plain CNN. Each column is an architecture. The layers with the same color have the same feature map size. Bold number indicates the number of filters.*

| CIFAR-10/100 | | | SVHN/ MNIST | |
|---|---|---|---|---|
| A | B | C | D | E |
| 9 layers | 9 layers | 7 layers | 7 layers | 7 layers |
| input image | | | | |
| conv3 -**64** | conv3 -**32** | conv3 -**32** | conv3 -**32** | conv3 -**16** |
| conv3 -**64** | conv3 -**32** | conv3 -**32** | conv3 -**32** | conv3 -**16** |
| conv3 -**128** | conv3 -**64** | conv3 -**64** | conv3 -**64** | conv3 -**32** |
| conv3 -**128** | conv3 -**64** | conv3 -**64** | conv3 -**64** | conv3 -**32** |
| conv3 -**256** | conv3 -**128** | conv3 -**128** | conv3-**128** | conv3 -**64** |
| conv3 -**256** | conv3 -**128** | conv3 -**128** | conv3-**128** | conv3 -**64** |
| conv3 -**256** | conv3 -**128** | | | |
| conv3 -**256** | conv3 -**128** | | | |
| global average pooling, FC-10/100, softmax | | | | |

An optional constraint to PCN was to use the same set of weights for both feedforward and feedback connections as in some prior studies (Rao and Ballard, 1999; Spratling, 2008, 2017). In other words, the weights of feedback connections were the transposed weights of feedforward connections. With this weight sharing, top-down predictions via feedback connections tended to approach lower-level representations. The PCN would have the same number of parameters as the corresponding plain model. Without this optional constraint of weight sharing, feedforward and feedback weights were assumed to be independent.

### 3.3 Recursive Computation

Unlike feedforward-only networks, PCN runs a dynamic process to update its internal representation throughout the hierarchy (Fig. 1.b). Given an input image, PCN first runs through the feedforward path from the input layer to the last convolutional layer at $\boldsymbol{t} = \mathbf{0}$, equivalent to a plain CNN model. For $\boldsymbol{t} = \mathbf{1}$, PCN first runs a feedback process and then a feedforward process to update the representations in the hierarchy. In the feedback process, the representation at each layer is updated by a top-down prediction from the higher layer according to Eq. (11). The feedback process runs from the highest convolutional layer to the input layer. In the feedforward process, the representation at each layer is updated by a bottom-up error according to Eq. (10). This procedure is repeated over time as shown in Fig. 1.b. After some cycles, the representation is used as the input to the classification layer to classify the image (see Algorithm 1).

### 3.4 Model Training

We evaluated two types of PCNs with regard to an optional constraint: the feedforward and feedback connections share the same convolutional weights. With this weight sharing, the feedforward operation and the feedback operation use the same weights. Without the constraint, the feedforward and feedback weights are initialized independently.

In this work, we evaluated these two types of PCNs with a varying number of recursive cycles ($t = 0, 1, 2, \cdots, 6$) and with different model architectures (labeled as A through E in Table 1). We use *Plain-A* to represent the plain network with architecture A, and use *PCN-A-t* to represent the PCN with architecture A and $t$ cycles of recursive computation. The numbers of recursive cycles for training and testing a model are the same. *PCN-A-t (tied)* and *PCN-A-t* represent the PCNs *with* and *without* weight sharing, respectively.

We used PyTorch (Paszke et al., 2017) to implement, train, and test the models described above. When PCN is trained for image classification, the classification error backpropagates across layers and in time to update the model parameters. The feedforward and feedback update rates ($a_l$ and $b_l$) are set to be specific to each filter in each layer, are constrained to be non-negative by using ReLU, and are trained with initial values $a_l = 1.0$ and $b_l = 0.5$, respectively. The convolutional weights and linear weights were initialized to be uniformly random (the default setting in PyTorch). The models were trained using mini-batches of a size 128 and without using dropout regularization (Srivastava et al., 2014).

**Algorithm 1** Deep Predictive Coding Network

Input static image: $\mathbf{x}$
$\mathbf{r}_0(t) \leftarrow \mathbf{x}$
*% initialize representations*
**for** $l = 0$ **to** L-1 **do**
  $\mathbf{r}_{l+1}(0) \leftarrow \text{ReLU}\left(\text{FFConv}\left(\mathbf{r}_l(0)\right)\right)$
*% recurrent computation with T cycles*
**for** t = 1 to T **do**
  *% nonlinear feedback process*
  **for** $l$ = L **to** 1 **do**
    $\mathbf{p}_{l-1}(t\text{-}1) \leftarrow \text{FBConv}\left(\mathbf{r}_l(t\text{-}1)\right)$
    **if** $l > 1$ **do**
      $\mathbf{r}_{l-1}(t\text{-}1) \leftarrow \text{ReLU}\left((1\text{-}b)\mathbf{r}_{l\text{-}1}(t\text{-}1) + b\mathbf{p}_{l\text{-}1}(t\text{-}1)\right)$
  *% nonlinear feedforward process*
  **for** $l$ = 0 **to** L-1 **do**
    $\mathbf{e}_l(t) \leftarrow \mathbf{r}_l(t) - \mathbf{p}_l(t\text{-}1)$
    $\mathbf{r}_{l+1}(t) \leftarrow \text{ReLU}\left(\mathbf{r}_{l+1}(t\text{-}1) + a\ \text{FFConv}\left(\mathbf{e}_l(t)\right)\right)$
*% classification*
Output $\mathbf{r}_\text{L}(\text{T})$ for classification

**Note:** *FFConv* represents the feedforward convolution, *FBConv* represents the feedback convolution. $a$ and $b$ are specific to each filter in each layer. *%comments* are comments

## 4. Experiments

We trained and tested PCN for image classification with data in CIFAR-10/100, SVHN and MNIST, in comparison with plain CNN using the same feedforward architecture. With random initialization, PCN (or CNN) was trained for 5 times; the best and mean±std top-1 accuracy was reported as below.

### 4.1 CIFAR-10 and CIFAR-100

The CIFAR-10/100 dataset includes 50,000 training images and 10,000 testing images in 10 or 100 object categories. Each image is a 32×32 RGB image. PCN (or CNN) were trained on the training set and evaluated on the test set. All images were normalized per channel (i.e. subtract the mean and divide by the standard deviation). For training, we used translation and horizontal flipping for data augmentation. We used stochastic gradient decent to train PCN (or CNN) with a weight decay of 0.0005 and a momentum of 0.9. The learning rate was initialized as 0.01 and was divided by 10 when the error reached the plateau after training for 80, 140, 200 epochs. We stopped after 250 epochs. The hyper-parameters for learning were set based on validation with 10,000 images in the training set.

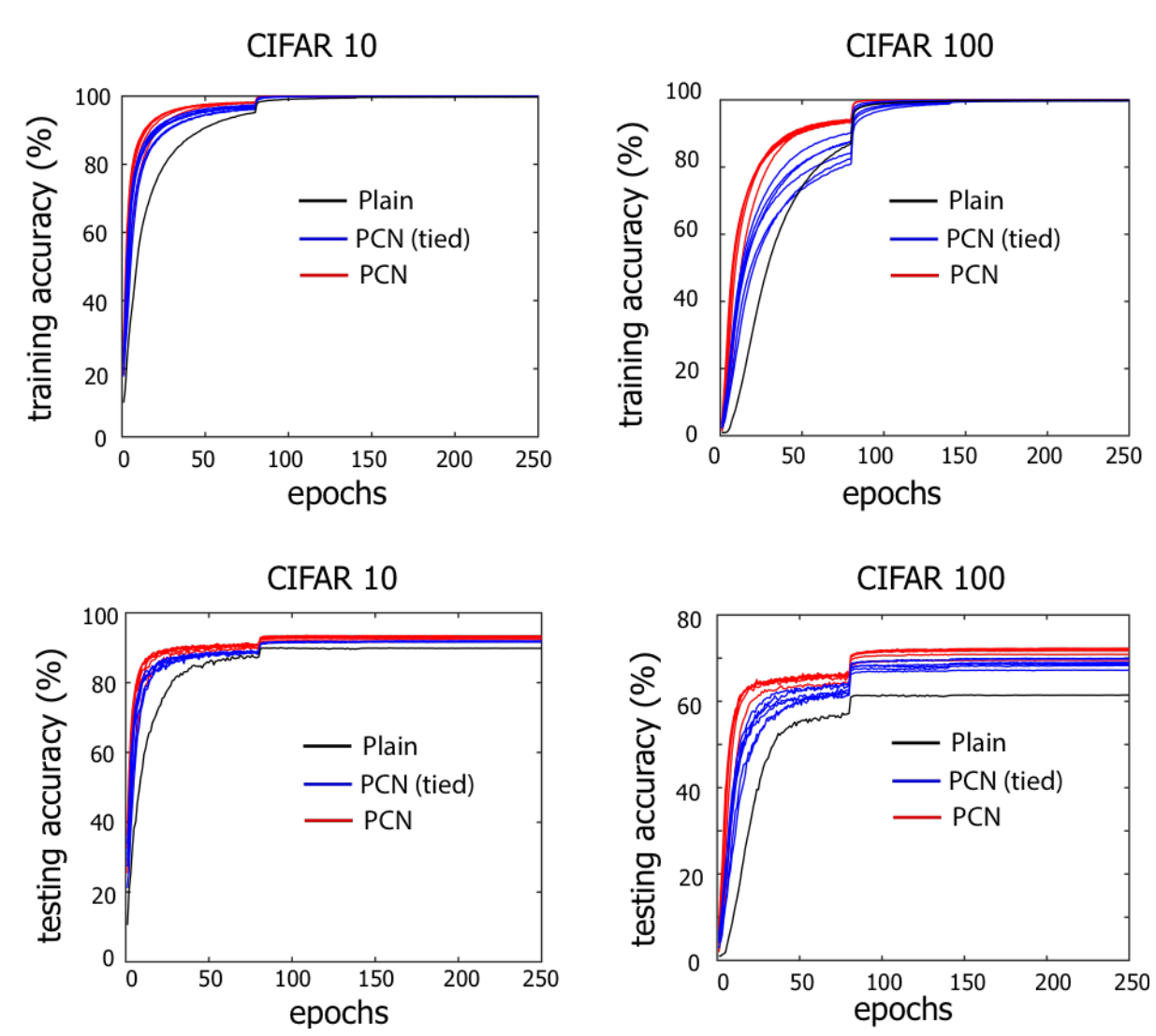

*Figure 2.* Training (top) and testing (bottom) accuracies for PCN vs. CNN with matched feedforward architectures for training with CIFAR-10 (left) and CIFAR-100 (right). Each curve represents the average over 5 repeats of one model with different cycles of recursive computation, ranging from 1 to 6.

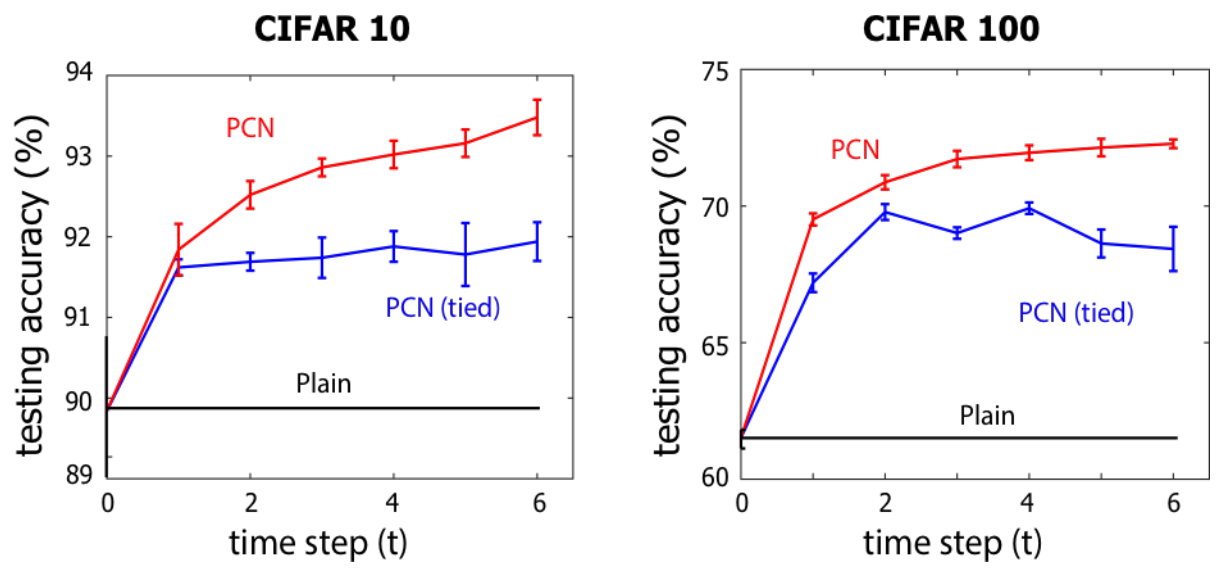

*Figure 3.* Testing accuracies of PCNs with different time steps.

### 4.1.1 PCN VS. CNN

During training, PCN converged much faster than its CNN counterpart (Fig. 2, top), especially when feedforward and feedback connections did not share weights. Meanwhile, increasing the recursive cycles tends to make PCN converge faster. With testing data, PCN also yielded better accuracy than the plain CNN model (Fig. 2, bottom). For example, *PCN improved* the classification accuracy from 62.11% to 72.48% on CIFAR-100, relative to the plain CNN model. See Table 2 for more results for comparison with other classical or state-of-the-art models. Without being pushed for high accuracy, PCN showed a similar accuracy as ResNet (He et al., 2016b), but relatively lower than the pre-activation ResNet (Pre-act-ResNet) (He et al., 2016a) or the wide residual network (WRN) (Zagoruyko and Komodakis, 2016), which used a much deeper or much wider architecture than the models explored in this study.

### 4.1.2 PCN WITH DIFFERENT RECURSIVE CYCLES

The accuracy of PCN depended on the number of cycles that recursively updated its internal representations. Fig. 3 shows that the accuracy of PCN tended to increase given more cycles of computation, especially if feedforward and feedback processes did not share the same weights.

To understand why this was the case, we looked into some testing images that were mis-classified by CNN but not by PCN. At each time step (0 through 6), PCN computed a different representation of an image that yielded a different probability distribution across different categories (Fig. 4). Classification was less definitive and/or inaccurate at early time steps. At later time steps, the network corrected itself to yield more definitive and accurate classification. It was true especially for ambiguous images, where a cat looked like a dog, or a deer looked like a horse, even for humans. See more examples in Fig. 4.

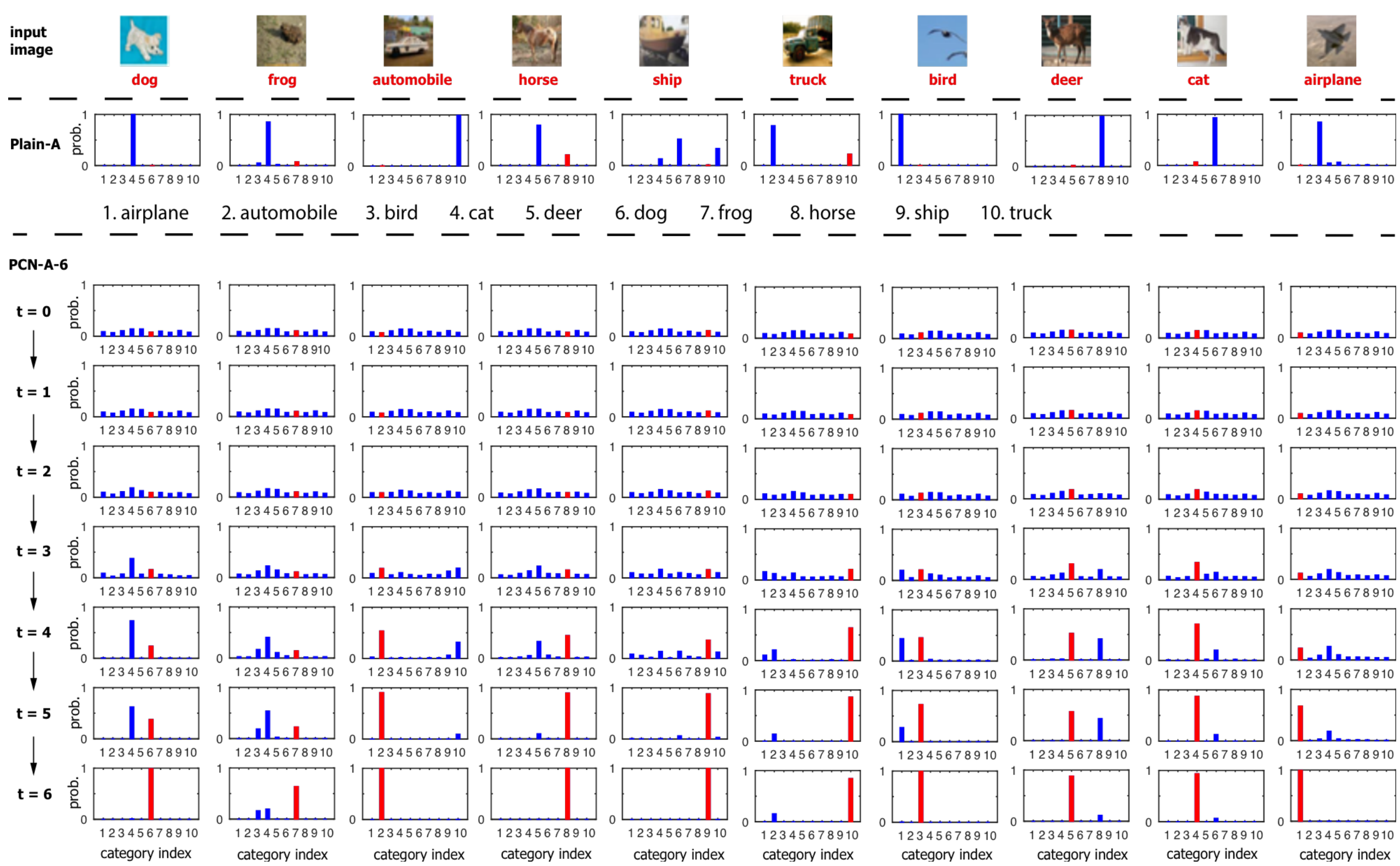

*Figure 4.* Image classification at different time steps for PCN-A-6 (bottom) in comparison with the plain CNN model (middle) for each of the 10 testing images misclassified by CNN (Plain-A). Each plot shows the probabilities over 10 classes in CIFAR-10. The red represents the ground truth.

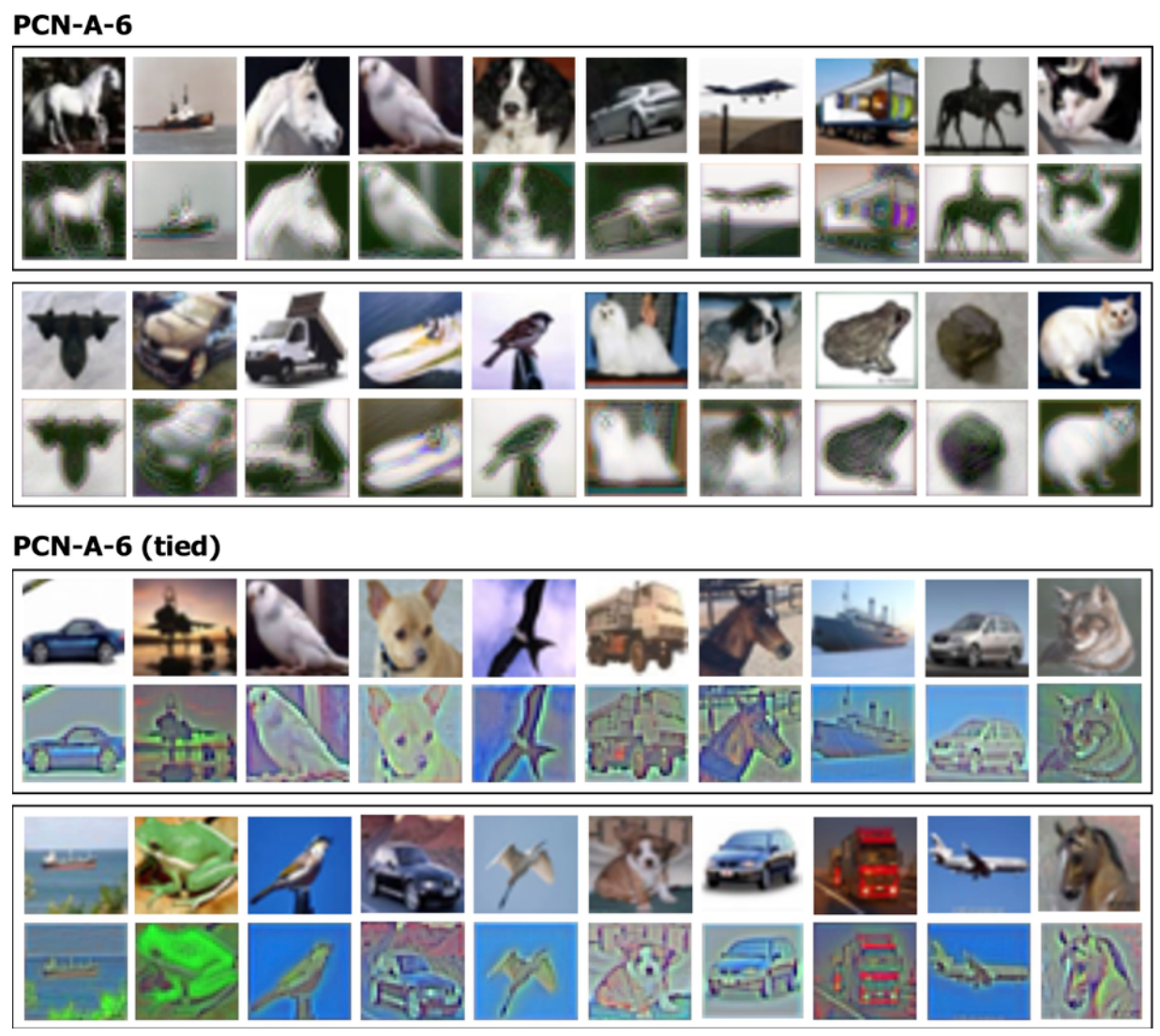

*Figure 5.* Top-down image prediction by PCN. Here shows example testing images in CIFAR-10 and their corresponding images predicted by PCNs.

*Table 2. Compare PCNs with start-of-the-art models on CIFAR-10/100 datasets. #L and #P are the number of layers and parameters, respectively.*

| Models | | | CIFAR10/100 | |
|---|---|---|---|---|
| Methods | #L | #P | Accuracy (%) | |
| Maxout(Goodfellow et al., 2013) | - | - | 90.62 | 61.43 |
| dasNet (Stollenga et al., 2014) | - | - | 90.78 | 66.22 |
| NIN (Lin et al., 2013) | - | - | 91.19 | 64.32 |
| DSN (Lee et al., 2015) | - | - | 91.78 | 65.43 |
| RCNN (Liang and Hu, 2015) | 6 | 1.86M | 92.91 | 68.25 |
| FitNet (Romero et al., 2014) | 19 | 2.5M | 91.61 | 64.96 |
| Highway(Srivastava et al., 2015) | 19 | 2.3M | 92.46 | 67.76 |
| ResNet (He et al., 2016b) | 110 | 1.7M | 93.57 | - |
| | 164 | 1.7M | - | 74.84 |
| | 1001 | 10.2M | - | 72.18 |
| | 1202 | 19.4M | 92.07 | - |
| Pre-act-ResNet (He et al., 2016a) | 110 | 1.7M | 93.63 | - |
| | 164 | 1.7M | 94.54 | 75.67 |
| | 1001 | 10.2M | 95.08 | 77.29 |
| WRN-40-4 | 40 | 8.9M | 95.47 | 78.82 |
| WRN-28-10 (Zagoruyko and Komodakis, 2016) | 28 | 36.5M | 96.00 | 80.75 |
| DenseNet (Huang et al., 2017) | 250 | 15.3M | 96.28 | 82.40 |
| **Plain-A** | 9 | 2.33M | 90.61 | 62.11 |
| **PCN-A-6 (tied)** | 9 | 2.33M | 92.26 | 69.44 |
| **PCN-A-6** | 9 | 4.65M | **93.83** | **72.58** |
| **Plain-B** | 9 | 0.58M | 89.53 | 62.21 |
| **PCN-B-2 (tied)** | 9 | 0.58M | 90.76 | 65.57 |
| **PCN-B-6** | 9 | 1.16M | 92.80 | 69.34 |
| **Plain-C** | 7 | 0.29M | 88.23 | 61.36 |
| **PCN-C-2 (tied)** | 7 | 0.29M | 89.56 | 64.09 |
| **PCN-C-6** | 7 | 0.57M | 92.40 | 68.31 |

#### 4.1.3 Generative Prediction in PCN

When it was trained for image classification, PCN was not explicitly optimized to reconstruct the input image, unlike a previous work that used video prediction as the learning objective (Lotter et al., 2016). Nevertheless, the top-down process in PCN was able to reconstruct the input with high accuracy. Although this was expected for PCN with weight sharing, reconstruction was also reasonable even for PCN without weight sharing (Fig. 5). This result was surprising, and implied that PCN, without any architectural constraint to enable image reconstruction, is able to reshape itself to predict or reconstruct the input, even when it is trained for a discriminative task, e.g. object recognition. Speculatively PCN potentially provides a new way to simultaneously train a discriminative network for object recognition and a generative network for prediction or reconstruction.

#### 4.1.4 Computational Requirement

Given a static image, CNN processes it with a single feedforward pass in testing, but PCN needs several cycles of recursive computation. For example, PCN-A-$t$ requires around $2t$ times the FLOPs of the plain CNN (0.68 billion FLOPs, multiply-adds). However, if the input is a video, CNN processes every video frame with a feedforward pass. PCN processes every frame with a feedback pass and a feedforward pass. Thus, PCN only doubles the FLOPs compared to the plain model given video input.

### 4.2 SVHN

SVHN is a dataset of Google's Street View House Numbers images (Netzer et al., 2011) and contains more than 600,000 color images of size 32×32, divided into training set, testing set and an extra set. The task of this dataset is to classify the digit located at the center of each image. Since the task is easier than CIFAR datasets, we implemented PCN with simpler network architectures (see Table 1). To validate the hyper parameters, we randomly selected 400 samples per class from the training set and 200 samples per class from the extra set for validation, as in (Goodfellow et al., 2013). The remainder of the training set and the extra set were used for training. The preprocessing for SVHN was the same as for CIFAR, i.e. per-channel normalization. No data augmentation was used. We used the Adam (Kingma and Ba, 2014) optimization with a weight decay of 0.0005 and an initial learning rate of 0.001 for a 20-10-10 epoch schedule. The exponential decay rates for the first and second moment estimates were 0.9 and 0.99, respectively. Table 3 shows the classification performance for this dataset. Like what we found for the CIFAR dataset, PCN always outperformed the plain CNN counterpart.

### 4.3 MNIST

The MNIST dataset consists of hand written digits 0-9. There are 60,000 training images and 10,000 testing images in total. Each image is a gray image of size 28x28. For this dataset, the same network architecture as used for SVHN is adopted. The training procedure was the same as for SVHN. Table 4 shows the classification performance for this dataset. PCN consistently performed better than its CNN counterpart. The best PCN achieves 0.36% error rate, comparable to some previous state-of-the-art models.

*Table 3. Compare PCNs with start-of-the-art models on SVHN. The accuracy was obtained from five repeats.*

| SVHN | | | |
|---|---|---|---|
| Methods | #L | #P | error rate (%) |
| Maxout(Goodfellow et al., 2013) | - | - | 2.47 |
| NIN (Lin et al., 2013) | - | - | 2.35 |
| Stochastic pooling (Zeiler and Fergus, 2013) | - | - | 2.80 |
| Dropconnect (Wan et al., 2013) | - | - | 1.94 |
| DSN (Lee et al., 2015) | - | - | 1.92 |
| RCNN (Liang and Hu, 2015) | 6 | 2.67M | 1.77 |
| FitNet (Romero et al., 2014) | 13 | 1.5M | 2.42 |
| WRN-16-8 (Zagoruyko and Komodakis, 2016) | 16 | 11M | 1.54 |
| **Plain-D** | 7 | 0.29M | 3.21(3.41±0.13) |
| **PCN-D-2 (tied)** | 7 | 0.29M | 2.63(2.92±0.11) |
| **PCN-D-6** | 7 | 0.57M | 2.28(2.42±0.09) |
| **Plain-E** | 7 | 0.07M | 3.19(3.41±0.13) |
| **PCN-E-1 (tied)** | 7 | 0.07M | 2.74(2.91±0.11) |
| **PCN-E-6** | 7 | 0.14M | **2.24**(2.42±0.10) |

*Table 4. Compare PCNs with the start-of-the-art models on MNIST. The accuracy was obtained from five repeats*

| MNIST | | | |
|---|---|---|---|
| Methods | #L | #P | error rate (%) |
| Maxout(Goodfellow et al., 2013) | - | - | 0.45 |
| NIN (Lin et al., 2013) | - | - | 0.47 |
| Stochastic pooling (Zeiler and Fergus, 2013) | - | - | 0.47 |
| Dropconnect (Wan et al., 2013) | - | - | 0.21 |
| DSN (Lee et al., 2015) | - | - | 0.39 |
| RCNN (Liang and Hu, 2015) | 6 | 0.67M | 0.31 |
| FitNet (Romero et al., 2014) | - | - | 0.51 |
| Hierarchical PC/BC-DIM (Spratling, 2017) | - | - | 2.19 |
| **Plain-D** | 7 | 0.29M | 0.53(0.59±0.04) |
| **PCN-D-1 (tied)** | 7 | 0.29M | 0.43(0.50±0.06) |
| **PCN-D-1** | 7 | 0.57M | 0.38(0.46±0.06) |
| **Plain-E** | 7 | 0.07M | 0.68(0.74±0.03) |
| **PCN-E-1 (tied)** | 7 | 0.07M | 0.43(0.51±0.06) |
| **PCN-E-4** | 7 | 0.14M | **0.36**(0.48±0.06) |

## 5. Discussion and Conclusion

What defines PCN are 1) the use of bi-directional and recurrent connections as opposed to feedforward-only connections, and 2) the use of predictive coding as a mechanism for the model to recursively run bottom-up and top-down processes. When it is trained for image classification, the model dynamically refines its representation of the input image towards more accurate and definitive recognition. As this computation is unfolded in time, PCN reuses a single architecture and the same set of parameters to run an increasingly longer cascade of nonlinear transformation.

We say it is "longer" instead of "deeper", because the notion behind PCN is different from the mindset in deep learning that more layers are required to model more complex and nonlinear relationships in data. In contrast, the brain does not use a deeper network to do more challenging tasks. A more challenging task simply takes the brain longer time to process information through the same network.

Predictive coding tells PCN how to compute but not how to learn. In this study, PCN is trained for image classification based on the representation emerging from the top layer after multiple cycles of computation. The error of classification backpropagates across layers and in time to update the model parameters per batch of training examples. This helps the learning to converge faster, while utilizing full knowledge in training data. If an image takes the model more cycles of computation to converge its representation, it means that the image has more information than what the model can explain or generate, and thus the image carries a greater value for the model to learn. Therefore, it is more desirable to train PCN for more challenging visual tasks, e.g. images that are ambiguous or difficult to recognize, while reducing the need for a large number of otherwise "simple" training examples.

For image classification, PCN takes an image as the input for all cycles of its recursive computation, while the errors of top-down prediction sent to the first hidden layer vary across cycles or in time. When the input is not a static image but a video, the input to the first hidden layer represents the errors of prediction of the present video frame given the model's representations from the past frames. This would enable the model to compute and learn representations of both spatial and temporal information in videos, which is an important aspect that awaits to be explored in future studies.

As an initial step to explore predictive coding in computer vision, it was our intention to start and compare with models with a basic CNN architecture (like that of VGG) in order to focus on evaluation of the value of using predictive coding as a computational mechanism. We expect that such predictive coding based computation can also be used to other network structures, e.g. ResNet and DenseNet. In a recent work (Han et al., 2018), a variant of PCN with a deeper structure and residual connections, has been developed and tested with ImageNet (Krizhevsky et al., 2012). It used notably fewer layers and parameters and but achieved competitive performance compared to classical and state-of-the-art models.

## Acknowledgments

The research was supported by NIH R01MH104402 and Purdue University.